\nonstopmode

\documentclass{article} 
\usepackage{nips15submit_e,times}
\usepackage[american]{babel}
\usepackage{csquotes}
\usepackage{afterpage}
\usepackage{amssymb,mathtools}
\usepackage{color,soul}
\usepackage{float}
\usepackage{graphicx}
\usepackage{hyperref}
\usepackage{nicefrac}
\usepackage{pgfplots}
\usepackage{subfig}
\usepackage{tabularx}
\usepackage{tikz}
\usepackage{url}

\newcommand*{\citep}{\cite}
\newcommand*{\citet}{\cite}

\title{Listen, Attend and Spell}

\author{
  William Chan \\
  Carnegie Mellon University \\
  \texttt{williamchan@cmu.edu} \\
  \And
  Navdeep Jaitly, Quoc V. Le, Oriol Vinyals \\
  Google Brain \\
  \texttt{\{ndjaitly,qvl,vinyals\}@google.com}
}

\nipsfinalcopy 

\begin{document}

\maketitle

\begin{abstract}





We present Listen, Attend and Spell (LAS), a neural network that learns to
transcribe speech utterances to characters. Unlike traditional DNN-HMM models,
this model learns all the components of a speech recognizer jointly.  Our
system has two components: a listener and a speller. The listener is a
pyramidal recurrent network encoder that accepts filter bank spectra as inputs.
The speller is an attention-based recurrent network decoder that emits
characters as outputs. The network produces character sequences without making
any independence assumptions between the characters. This is the key
improvement of LAS over previous end-to-end CTC models.  On a subset of the Google voice
search task, LAS achieves a word error rate (WER) of 14.1\% without a
dictionary or a language model, and 10.3\% with language model rescoring over
the top 32 beams.  By comparison, the state-of-the-art CLDNN-HMM model achieves
a WER of 8.0\%.

\end{abstract}

\section{Introduction}
Deep Neural Networks (DNNs) have led to improvements in various components of
speech recognizers. They are commonly used in hybrid DNN-HMM speech recognition
systems for acoustic modeling
~\citep{morgan-icassp-1990,mohamed2009,dahl2011large,mohamed2012acoustic,jaitly-interspeech-2012,
sainath-icassp-2013}. DNNs have also produced significant gains in
pronunciation models that map words to phoneme sequences
\citep{rao2015grapheme,yao2015}. In language modeling, recurrent models have
been shown to improve speech recognition accuracy by rescoring n-best lists
\citep{mikolov-interspeech-2010}. Traditionally these components -- acoustic,
pronunciation and language models -- have all been trained separately, each
with a different objective. Recent work in this area attempts to rectify this
disjoint training issue by designing models that are trained end-to-end -- from
speech directly to
transcripts~\citep{graves-icml-2006,graves-icml-2012,graves-icml-2014,awni-arxiv-2014,chorowski-nips-2014,chorowski-nips-2015}.
Two main approaches for this are Connectionist Temporal Classification (CTC)
\citep{graves-icml-2006} and sequence to sequence models with
attention~\citep{bahdanau-iclr-2015}.  Both of these approaches have limitations
that we try to address: CTC assumes that the label outputs are conditionally
independent of each other; whereas the sequence to sequence approach has only
been applied to phoneme sequences
\citep{chorowski-nips-2014,chorowski-nips-2015}, and not trained end-to-end for
speech recognition.

In this paper we introduce Listen, Attend and Spell (LAS), a neural network
that improves upon the previous
attempts~\citep{graves-icml-2014,chorowski-nips-2014,chorowski-nips-2015}.  The
network learns to transcribe an audio sequence signal to a word sequence, one
character at a time. Unlike previous approaches, LAS does not make
independence assumptions in the label sequence and it does not rely on HMMs.
LAS is based on the sequence to sequence learning
framework with attention~\citep{sutskever-nips-2014,cho-emnlp-2014,bahdanau-iclr-2015,chorowski-nips-2014,chorowski-nips-2015}.
It consists of an encoder recurrent neural network (RNN), which is named the
\emph{listener}, and a decoder RNN, which is named the \emph{speller}. The
listener is a pyramidal RNN that converts low level speech signals into higher
level features. The speller is an RNN that converts these higher level features
into output utterances by specifying a probability distribution over sequences
of characters using the attention
mechanism~\citep{bahdanau-iclr-2015,chorowski-nips-2014,chorowski-nips-2015}.
The listener and the speller are trained jointly.

Key to our approach is the fact that we use a pyramidal RNN model for the
listener, which reduces the number of time steps that the attention model has
to extract relevant information from. Rare and out-of-vocabulary (OOV) words
are handled automatically, since the model outputs the character sequence, one
character at a time. Another advantage of modeling characters as outputs is
that the network is able to generate multiple spelling variants naturally. For
example, for the phrase ``triple a'' the model produces both ``triple a''
and ``aaa'' in the top beams (see section~\ref{sec:int_examples}). A model like
CTC may have trouble producing such diverse transcripts for the same utterance
because of conditional independence assumptions between frames.

In our experiments, we find that these components are necessary for
LAS to work well. Without the attention mechanism, the model
overfits the training data significantly, in spite of our large
training set of three million utterances - it memorizes the training
transcripts without paying attention to the acoustics. Without the
pyramid structure in the encoder side, our model converges too slowly
- even after a month of training, the error rates were significantly
higher than the errors we report here. Both of these problems arise
because the acoustic signals can have hundreds to thousands of frames
which makes it difficult to train the RNNs. Finally, to reduce the
overfitting of the speller to the training transcripts, we use a
sampling trick during training~\citep{bengio-arxiv-2015}.

With these improvements, LAS achieves 14.1\% WER on a subset of the Google voice
search task, without a dictionary or a language model. When combined
with language model rescoring, LAS achieves 10.3\% WER. By
comparison, the Google state-of-the-art CLDNN-HMM system achieves 8.0\% WER on the
same data set~\citep{sainath-icassp-2015}.

\section{Related Work}
Even though deep networks have been successfully used in many
applications, until recently, they have mainly been used in
classification: mapping a fixed-length vector to an output
category~\cite{krizhevsky-nips-2012}. For structured problems, such as
mapping one variable-length sequence to another variable-length
sequence, neural networks have to be combined with other sequential
models such as Hidden Markov Models (HMMs)~\cite{baum1966statistical}
and Conditional Random Fields (CRFs) \cite{Lafferty2001}.  A drawback
of this combining approach is that the resulting models cannot be
easily trained end-to-end and they make simplistic assumptions about
the probability distribution of the data.

Sequence to sequence learning is a framework that attempts to address
the problem of learning variable-length input and output
sequences~\cite{sutskever-nips-2014}. It uses an encoder RNN to map
the sequential variable-length input into a fixed-length vector. A
decoder RNN then uses this vector to produce the variable-length
output sequence, one token at a time. During training, the model feeds
the groundtruth labels as inputs to the decoder. During inference, the
model performs a beam search to generate suitable candidates for next
step predictions.

Sequence to sequence models can be improved significantly by the use of
an attention mechanism that provides the decoder RNN more information
when it produces the output tokens~\cite{bahdanau-iclr-2015}. At each
output step, the last hidden state of the decoder RNN is used to generate
an attention vector over the input sequence of the encoder. The attention
vector is used to propagate information from the encoder to the decoder
at every time step, instead of just once, as with the original sequence to
sequence model~\cite{sutskever-nips-2014}. This attention vector can be
thought of as skip connections that allow the information and the gradients
to flow more effectively in an RNN.

The sequence to sequence framework has been used extensively for many
applications: machine translation~\cite{luong-acl-2015,jean-acl-2015}, image
captioning~\cite{vinyals-arvix-2014,xu-icml-2015}, parsing~\cite{vinyals-2014}
and conversational modeling~\cite{vinyals-icml-2015}. The generality of this
framework suggests that speech recognition can also be a direct application
\cite{chorowski-nips-2014,chorowski-nips-2015}.

\section{Model}
In this section, we will formally describe LAS which accepts acoustic
features as inputs and emits English characters as outputs. Let $\mathbf x =
(x_1, \dots, x_T)$ be our input sequence of filter bank spectra features, and
let $\mathbf y = (\langle \mathrm{sos} \rangle, y_1, \dots, y_S, \langle
\mathrm{eos} \rangle)$, $y_i \in \{a, b, c, \cdots, z, 0, \cdots, 9, \langle
\mathrm{space} \rangle, \langle \mathrm{comma} \rangle, \langle \mathrm{period}
\rangle, \langle \mathrm{apostrophe} \rangle, \langle \mathrm{unk} \rangle \}$,
be the output sequence of characters.  Here $\langle \mathrm{sos} \rangle$ and
$\langle \mathrm{eos} \rangle$ are the special start-of-sentence token, and
end-of-sentence tokens, respectively.

We want to model each character output $y_i$ as a conditional distribution
over the previous characters $y_{<i}$ and the input signal $\mathbf x$
using the chain rule:
\begin{align}
  P(\mathbf y | \mathbf x) = \prod_i P(y_i | \mathbf x, y_{<i})
\end{align}

Our Listen, Attend and Spell (LAS) model consists of two
sub-modules: the listener and the speller. The listener is an acoustic model
encoder, whose key operation is $\operatorname{Listen}$. The speller is an
attention-based character decoder, whose key operation is
$\operatorname{AttendAndSpell}$. The $\operatorname{Listen}$ function
transforms the original signal $\mathbf x$ into a high level representation
$\mathbf h = (h_1, \dots, h_U)$ with $U \le T$, while the
$\operatorname{AttendAndSpell}$ function consumes $\mathbf h$ and produces a
probability distribution over character sequences: \begin{align} \mathbf h &=
\operatorname{Listen}(\mathbf x) \\ P(\mathbf y | \mathbf x) &=
\operatorname{AttendAndSpell}(\mathbf h, \mathbf y) \end{align}

\begin{figure}[h!] \centering \resizebox{0.8\textwidth}{!}{\input{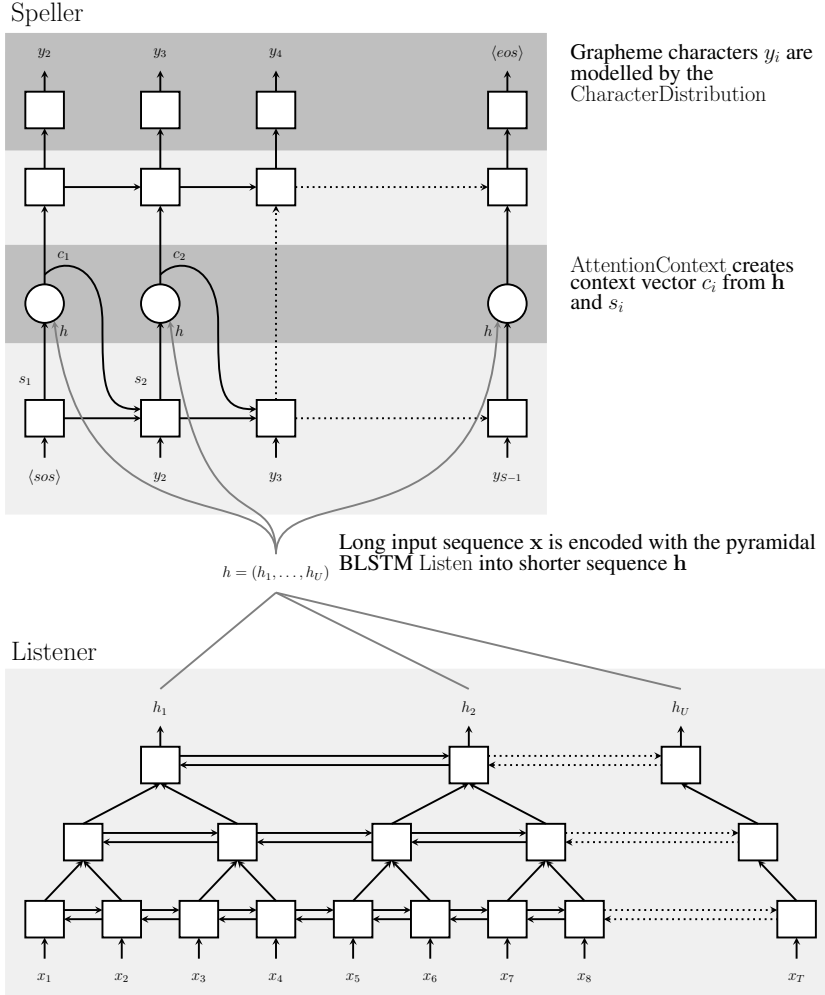}}
  \caption{\small Listen, Attend and Spell (LAS) model: the listener is a
  pyramidal BLSTM encoding our input sequence $\mathbf x$ into high level features $\mathbf h$,
the speller is an attention-based decoder generating the $\mathbf y$ characters from $\mathbf h$.} 
\label{fig:model} 
\end{figure}

Figure~\ref{fig:model} visualizes LAS with these two components. We
provide more details of these components in the following sections.

\subsection{Listen}
The $\operatorname{Listen}$ operation uses a Bidirectional Long Short
Term Memory RNN (BLSTM)
\cite{hochreiter-neuralcomputation-1997,graves-asru-2013,graves-icml-2014}
with a pyramid structure. This modification is required to reduce the
length $U$ of ${\mathbf h}$, from $T$, the length of the input
${\mathbf x}$, because the input speech signals can be hundreds to
thousands of frames long.  A direct application of BLSTM for the
operation $\operatorname{Listen}$ converged slowly and produced
results inferior to those reported here, even after a month of
training time.  This is presumably because the operation
$\operatorname{AttendAndSpell}$ has a hard time extracting the
relevant information from a large number of input time steps.

We circumvent this problem by using a pyramid BLSTM (pBLSTM) similar
to the Clockwork RNN \cite{koutnik-icml-2014}. In each successive stacked pBLSTM
layer, we reduce the time resolution by a factor of 2. In a typical deep
BTLM architecture, the output at the $i$-th time step, from the $j$-th layer is
computed as follows:
\begin{align}
  h_i^{j} = \operatorname{BLSTM}(h_{i - 1}^{j}, h_{i}^{j - 1})
\end{align}
In the pBLSTM model, we concatenate the outputs at consecutive
steps of each layer before feeding it to the next layer, i.e.:
\begin{align}
  h_i^{j} = \operatorname{pBLSTM}(h_{i - 1}^{j}, \left[ h_{2i}^{j - 1}, h_{2i + 1}^{j - 1} \right])
\end{align}

In our model, we stack 3 pBLSTMs on top of the bottom BLSTM layer to
reduce the time resolution $2^3=8$ times. This allows the attention
model (see next section) to extract the relevant information from a
smaller number of times steps. In addition to reducing the resolution,
the deep architecture allows the model to learn nonlinear feature
representations of the data. See Figure \ref{fig:model} for a
visualization of the pBLSTM.

The pyramid structure also reduces the computational complexity. In
the next section we show that the attention mechanism over $U$
features has a computational complexity of $O(US)$. Thus,
reducing $U$ speeds up learning and inference significantly.

\subsection{Attend and Spell}
We now describe the $\operatorname{AttendAndSpell}$ function. The
function is computed using an attention-based LSTM transducer
\cite{bahdanau-iclr-2015,chorowski-nips-2015}.
At every output step, the transducer produces a probability distribution
over the next character conditioned on all the characters seen previously.
The distribution for $y_i$ is a function of the decoder state $s_i$ and
context $c_i$. The decoder state $s_i$ is a function of the previous
state $s_{i - 1}$, the previously emitted character $y_{i - 1}$
and context $c_{i - 1}$. The context vector $c_i$ is produced by an
attention mechanism. Specifically,
\begin{align}
  c_i &= \operatorname{AttentionContext}(s_i, \mathbf h) \\
  s_i &= \operatorname{RNN}(s_{i - 1}, y_{i - 1}, c_{i - 1}) \\
  P(y_i | \mathbf x, y_{<i}) &= \operatorname{CharacterDistribution}(s_i, c_i)
\end{align} where
$\operatorname{CharacterDistribution}$ is an MLP with softmax outputs over
characters, and $\operatorname{RNN}$ is a 2 layer LSTM.

At each time step, $i$, the attention mechanism, $\operatorname{AttentionContext}$
generates a context vector, $c_i$ encapsulating the information in the acoustic
signal needed to generate the next character.  The attention model is
content based - the contents of the decoder state $s_i$ are matched
to the contents of $h_u$ representing time step $u$ of $\mathbf h$, to generate an
attention vector $\alpha_i$. $\alpha_i$ is used to linearly blend vectors $h_u$ to
create $c_i$.

Specifically, at each decoder timestep $i$, the $\operatorname{AttentionContext}$
function computes the scalar energy $e_{i, u}$ for each time step $u$, using
vector $h_u \in \mathbf h$ and $s_i$. The scalar energy $e_{i,u}$ is converted into a
probability distribution over times steps (or attention) $\alpha_i$ using a 
softmax function. This is used to create the context vector $c_i$ by linearly
blending the listener features, $h_u$, at different time steps:
\begin{align}
  e_{i,  u}      &= \langle \phi(s_i), \psi(h_u) \rangle \\
  \alpha_{i, u} &= \frac{\exp(e_{i, u})}{\sum_u \exp(e_{i, u})} \\
  c_i                   &= \sum_u \alpha_{i, u} h_u \end{align}
where
$\phi$ and $\psi$ are MLP networks. On convergence, the $\alpha_{i}$
distribution is typically very sharp, and focused on only a few frames of $\mathbf h$;
$c_i$ can be seen as a continuous bag of weighted features of $\mathbf h$. Figure
\ref{fig:model} shows LAS architecture.

\subsection{Learning}
\label{sec:learning}
The $\operatorname{Listen}$ and $\operatorname{AttendAndSpell}$ functions can
be trained jointly for end-to-end speech recognition. The sequence to sequence
methods condition the next step prediction on the previous
characters~\cite{sutskever-nips-2014,bahdanau-iclr-2015} and maximizes the log
probability:
\begin{align}
  \max_\theta \sum_i \log P(y_i | \mathbf x, y^*_{<i};
\theta)
\end{align} where $y^*_{<i}$ is the groundtruth of the previous
characters.

However during inference, the groundtruth is missing and the predictions can suffer
because the model was not trained to be resilient to feeding in bad predictions
at some time steps. To ameliorate this effect, we use a trick that was proposed in \cite{bengio-arxiv-2015}.
During training, instead of always feeding in the ground truth transcript for next step
prediction, we sometimes sample from our previous character distribution and use that as the
inputs in the next step predictions:
\begin{gather}
  \tilde{y}_{i} \sim \operatorname{CharacterDistribution}(s_{i}, c_{i}) \\
  \max_\theta \sum_i \log P(y_i | \mathbf x, \tilde{y}_{< i}; \theta)
\end{gather}
where $\tilde{y}_{i - 1}$ is the character chosen from the ground truth, or sampled from
the model with a certain sampling rate. Unlike \cite{bengio-arxiv-2015}, we do not use
a schedule and simply use a constant sampling rate of 10\% right from the start
of training.

As the system is a very deep network it may appear that some type of
pretraining would be required. However, in our experiments, we found no need
for pretraining. In particular, we attempted to pretrain the
$\operatorname{Listen}$ function with context independent or context dependent
phonemes generated from a conventional GMM-HMM system. A softmax network was
attached to the output units $h_u \in \mathbf h$ of the listener and used to
make multi-frame phoneme state predictions \cite{navdeep-interspeech-2014} but
led to no improvements. We also attempted to use the phonemes as a joint
objective target \cite{sak-interspeech-2015}, but found no improvements.

\subsection{Decoding and Rescoring}
During inference we want to find the most likely character sequence given
the input acoustics:
\begin{align}
  \hat{\mathbf y} = \arg \max_{\mathbf y} \log P(\mathbf y | \mathbf x)
\end{align}
Decoding is performed with a simple left-to-right beam search algorithm similar
to \cite{sutskever-nips-2014}. We maintain a set of $\beta$ partial
hypotheses, starting with the start-of-sentence $\langle \mathrm{sos}
\rangle$ token. At each timestep, each partial hypothesis in the
beam is expanded with every possible character and only the
$\beta$ most likely beams are kept. When the $\langle \mathrm{eos} \rangle$
token is encountered, it is removed from the beam and added to the set of
complete hypothesis. A dictionary can optionally be added to constrain
the search space to valid words, however we found that this was not necessary
since the model learns to spell real words almost all the time.

We have vast quantities of text data \cite{mikolov-nips-2013}, compared to
the amount of transcribed speech utterances. We can use language models
trained on text corpora alone similar to conventional speech systems \cite{povey-asru-2011}.
To do so we can rescore our beams with the language model. We find that our model
has a small bias for shorter utterances so we normalize our probabilities by the
number of characters $|\mathbf y|_c$ in the hypothesis and combine it with a language model
probability $P_{\operatorname{LM}}(\mathbf y)$: \begin{align} s(\mathbf y |
\mathbf x) &= \frac{\log P(\mathbf y | \mathbf x)}{|\mathbf y|_c} + \lambda
\log P_{\operatorname{LM}}(\mathbf y) \end{align} where $\lambda$ is our
language model weight and can be determined by a held-out validation set.

\section{Experiments}
\makeatletter
\let\percent\@percentchar
\makeatother

\label{sec:experiments}
We used a dataset approximately three million Google voice search utterances
(representing 2000 hours of data) for our experiments.  Approximately 10 hours
of utterances were randomly selected as a held-out validation set. Data
augmentation was performed using a room simulator, adding different types of
noise and reverberations; the noise sources were obtained from YouTube and
environmental recordings of daily events \citep{sainath-icassp-2015}.  This
increased the amount of audio data by $20$ times. 40-dimensional log-mel filter
bank features were computed every 10ms and used as the acoustic inputs to the
listener.  A separate set of 22K utterances representing approximately 16 hours
of data were used as the test data. A noisy test data set was also created
using the same corruption strategy that was applied to the training data. All
training sets are anonymized and hand-transcribed, and are representative of
Google’s speech traffic.


The text was normalized by converting all characters to lower case
English alphanumerics (including digits). The punctuations: space,
comma, period and apostrophe were kept, while all other tokens were
converted to the unknown $\langle \mathrm{unk} \rangle$ token. As
mentioned earlier, all utterances were padded with the
start-of-sentence $\langle \mathrm{sos} \rangle$ and the
end-of-sentence $\langle \mathrm{eos} \rangle$ tokens.

The state-of-the-art model on this dataset is a CLDNN-HMM system that
was described in \citep{sainath-icassp-2015}. The CLDNN system achieves
a WER of 8.0\% on the clean test set and 8.9\% on the noisy test
set. However, we note that the CLDNN uses unidirectional CLDNNs and
would certainly benefit even further from the use of a bidirectional
CLDNN architecture.

For the $\operatorname{Listen}$ function we used 3 layers of 512 pBLSTM nodes
(i.e., 256 nodes per direction) on top of a BLSTM that operates on the input.
This reduced the time resolution by $8 = 2^3$ times. The $\operatorname{Spell}$
function used a two layer LSTM with 512 nodes each. The weights were
initialized with a uniform distribution $\mathcal{U}(-0.1, 0.1)$.

Asynchronous Stochastic Gradient Descent (ASGD) was used for training
our model~\citep{dean-nips-2012}. A learning rate of $0.2$ was used
with a geometric decay of $0.98$ per 3M utterances
(i.e., $\nicefrac{1}{20}$-th of an epoch). We used the DistBelief
framework~\citep{dean-nips-2012} with 32 replicas, each with a
minibatch of 32 utterances. In order to further speed up training, the
sequences were grouped into buckets based on their frame
length~\citep{sutskever-nips-2014}.

The model was trained using groundtruth previous characters until results on the
validation set stopped improving. This took approximately two weeks. The model
was decoded using beam width $\beta=32$ and achieved 16.2\% WER on
the clean test set and 19.0\% WER on the noisy test set without any
dictionary or language model. We found that constraining the beam search with a
dictionary had no impact on the WER. Rescoring the top 32 beams with
the same n-gram language model that was used by the CLDNN system using a language model
weight of $\lambda=0.008$ improved the results for the clean
and noisy test sets to 12.6\% and 14.7\% respectively.  Note that for convenience,
we did not decode with a language model, but rather only rescored the top 32 beams.
It is possible that further gains could have been achieved by using the language model
during decoding.

\begin{table}[t]
  \centering
  \caption{\small WER comparison on the clean and noisy Google voice search
    task. The CLDNN-HMM system is the state-of-the-art system, the
    Listen, Attend and Spell (LAS) models are decoded with a beam size
    of 32. Language Model (LM) rescoring was applied to our beams, and a
    sampling trick was applied to bridge the gap between training and
    inference.}
  \label{tab:wer}
  \begin{tabular}{|l|l|l|}
    \hline
    \bfseries Model & \bfseries Clean WER & \bfseries Noisy WER \\
    \hline
    \hline
    CLDNN-HMM \citep{sainath-icassp-2015} & 8.0 & 8.9 \\
    \hline
    LAS                           & 16.2 & 19.0 \\
    LAS + LM Rescoring            & 12.6 & 14.7 \\
    LAS + Sampling                & 14.1 & 16.5 \\
    LAS + Sampling + LM Rescoring & 10.3 & 12.0 \\
    \hline
  \end{tabular}
\end{table}

As mentioned in Section \ref{sec:learning}, there is a mismatch between
training and testing. During training the model is conditioned on the correct
previous characters but during testing mistakes made by the model corrupt
future predictions. We trained another model by sampling from our previous character
distribution with a probability of 10\% (we did not use a schedule as
described in~\citep{bengio-arxiv-2015}).  This improved our results on the clean
and noisy test sets to 14.1\% and 16.5\% WER respectively when no language model rescoring
was used. With language model rescoring, we achevied 10.3\% and 12.0\% WER on the clean
and noisy test sets, respectively. Table \ref{tab:wer} summarizes these results.

On the clean test set, this model is within 2.5\% absolute WER of the
state-of-the-art CLDNN-HMM system, while on the noisy set it is less than
3.0\% absolute WER worse. We suspect that convolutional filters could lead to
improved results, as they have been reported to improve performance by 5\%
relative WER on clean speech and 7\% relative on noisy speech compared to
non-convolutional architectures \citep{sainath-icassp-2015}.

\subsection{Attention Visualization}
\begin{figure}[h!]
  \centering
  \includegraphics[width=4.5in]{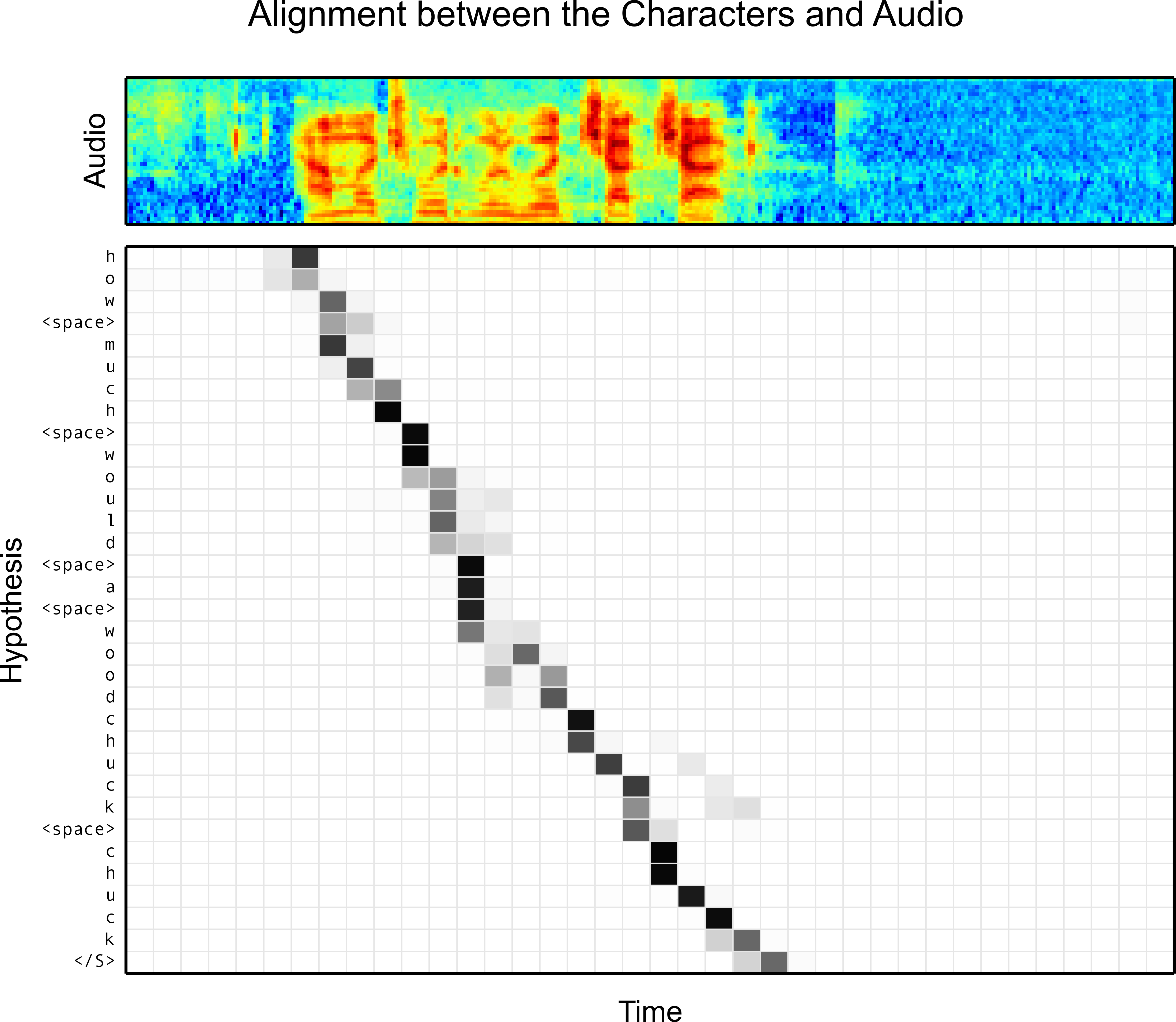}
  \caption{\small Alignments between character outputs and audio signal produced by the Listen, Attend and Spell (LAS) model for the utterance ``how much would a woodchuck chuck''. The content based attention mechanism was able to identify the start position in the audio sequence for the first character correctly. The alignment produced is generally monotonic without a need for any location based priors.}
  \label{fig:woodchuck}
\end{figure}
The content-based attention mechanism creates an explicit alignment between the
characters and audio signal. We can visualize the attention mechanism by
recording the attention distribution on the acoustic sequence at every
character output timestep. Figure \ref{fig:woodchuck} visualizes the attention
alignment between the characters and the filterbanks for the utterance ``how
much would a woodchuck chuck''. For this particular utterance, the model learnt
a monotonic distribution without any location priors. The words ``woodchuck''
and ``chuck'' have acoustic similarities, the attention mechanism was slightly
confused when emitting ``woodchuck'' with a dilution in the distribution. The
attention model was also able to identify the start and end of the utterance
properly.

In the following sections, we report results of control experiments
that were conducted to understand the effects of beam widths,
utterance lengths and word frequency on the WER of our model.

\subsection{Effects of Beam Width}
\begin{figure}[h!] \centering
  \includegraphics{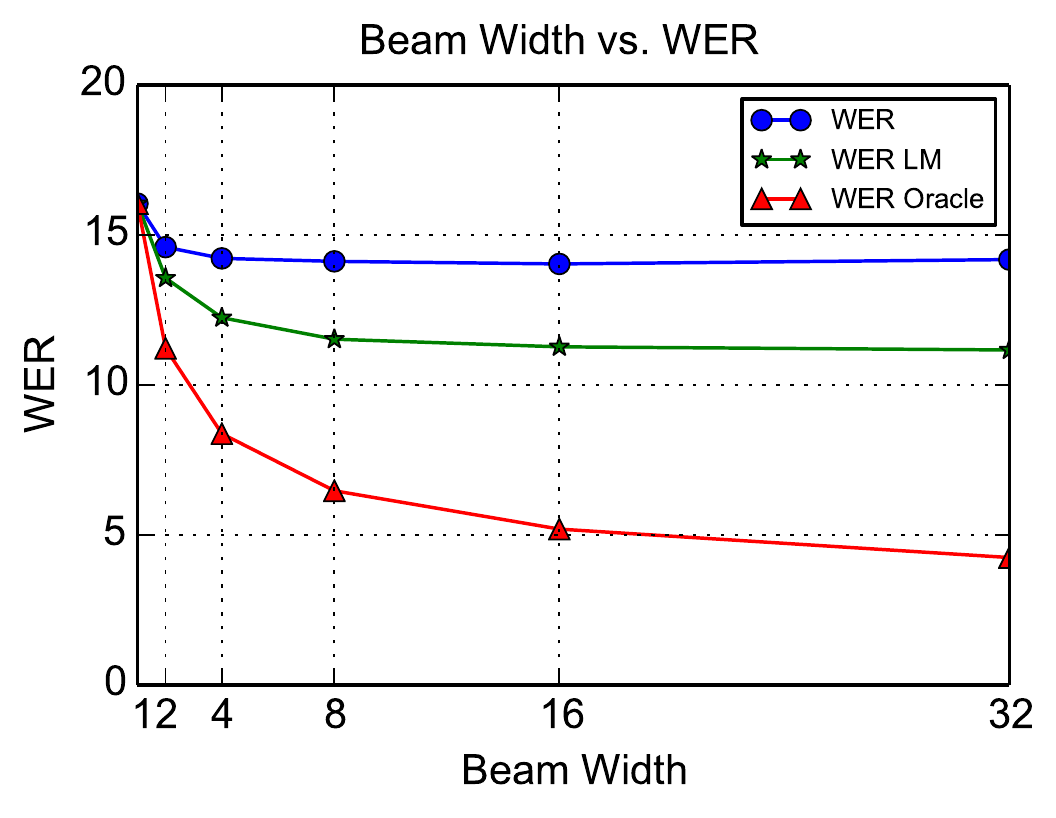} \caption{\small The effect of the
    decode beam width on WER for the clean Google voice search
    task. The reported WERs are without a dictionary or language
    model, with language model rescoring and the oracle WER for
    different beam widths. The figure shows that good results can be
    obtained even with a relatively small beam
    size.} \label{fig:beamwidth}
\end{figure}

We investigate the correlation between the performance of the model
and the width of beam search, with and without the language model
rescoring. Figure~\ref{fig:beamwidth} shows the effect of the decode
beam width, $\beta$, on the WER for the clean test set.  We see
consistent WER improvements by increasing the beam width up to 16,
after which we observe no significant benefits. At a beam width of 32,
the WER is 14.1\% and 10.3\% after language model rescoring.
Rescoring the top 32 beams with an oracle produces a WER of 4.3\% on
the clean test set and 5.5\% on the noisy test
set.  

\subsection{Effects of Utterance Length}
We measure the performance of our model as a function of the number of
words in the utterance. We expect the model to do poorly on longer
utterances due to limited number of long training utterances in our
distribution. Hence it is not surprising that longer utterances have a
larger error rate. The deletions dominate the error for long
utterances, suggesting we may be missing out on words. It is
surprising that short utterances (e.g., 2 words or less) perform quite
poorly. Here, the substitutions and insertions are the main sources of
errors, suggesting the model may split words apart.

\begin{figure}[h!] \centering
  \includegraphics{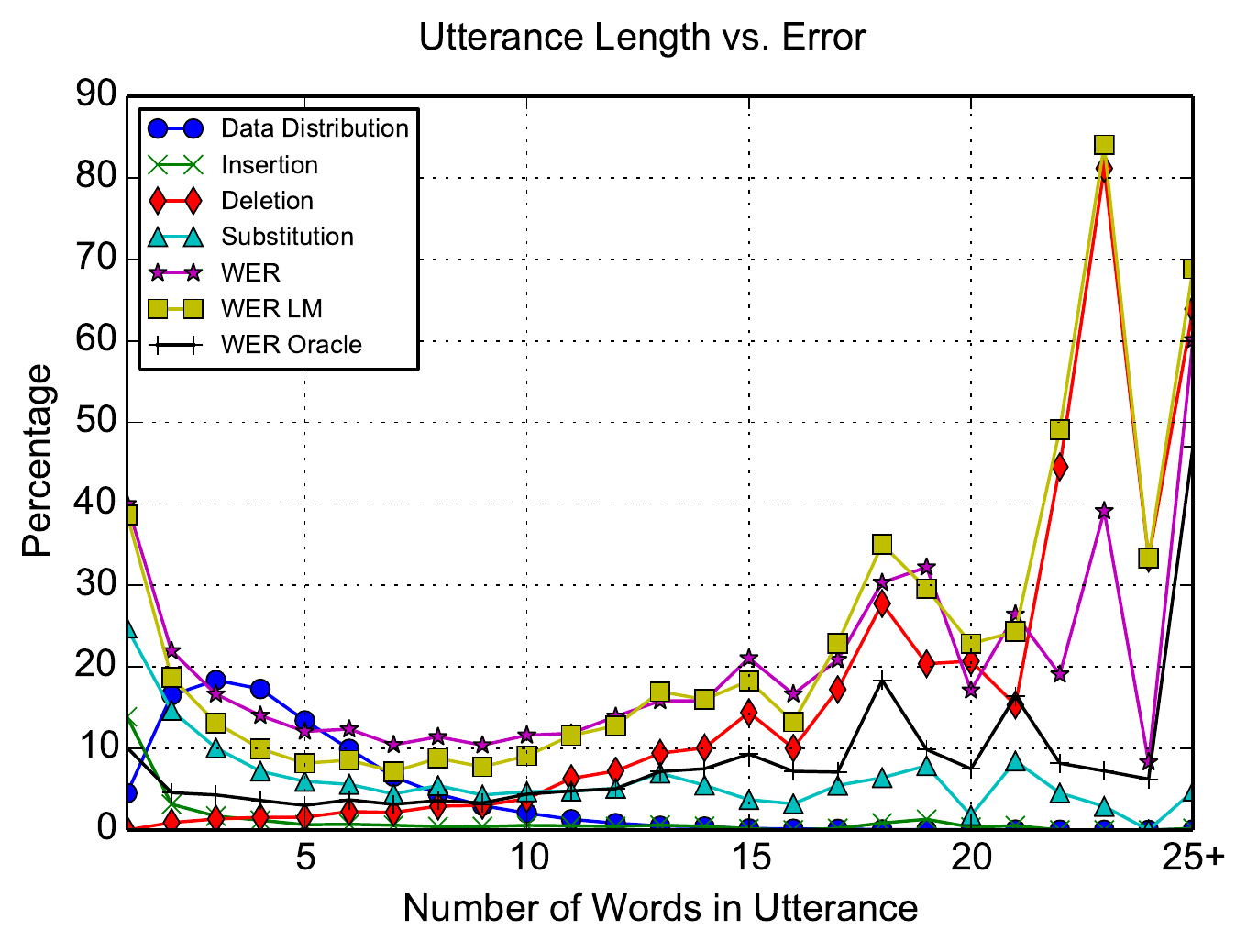} \caption{\small The correlation between
    error rates (insertion, deletion, substitution and WER) and the number of
    words in an utterance. The WER is reported without a dictionary or language
    model, with language model rescoring and the oracle WER for the clean
    Google voice search task. The data distribution with respect to the number
    of words in an utterance is overlaid in the figure. LAS performs
    poorly with short utterances despite an abundance of data. LAS also
    fails to generalize well on longer utterances when trained on a distribution
    of shorter utterances. Insertions and substitutions are the main sources of errors for
    short utterances, while deletions dominate the error for long utterances.} \label{fig:wordlength} \end{figure}

Figure \ref{fig:wordlength} also suggests that our model struggles to
generalize to long utterances when trained on a distribution of
shorter utterances. It is possible location-based priors may help in
these situations as reported by~\citep{chorowski-nips-2015}.

\subsection{Word Frequency}
\begin{figure}[h!] \centering
  \includegraphics[width=0.8\textwidth]{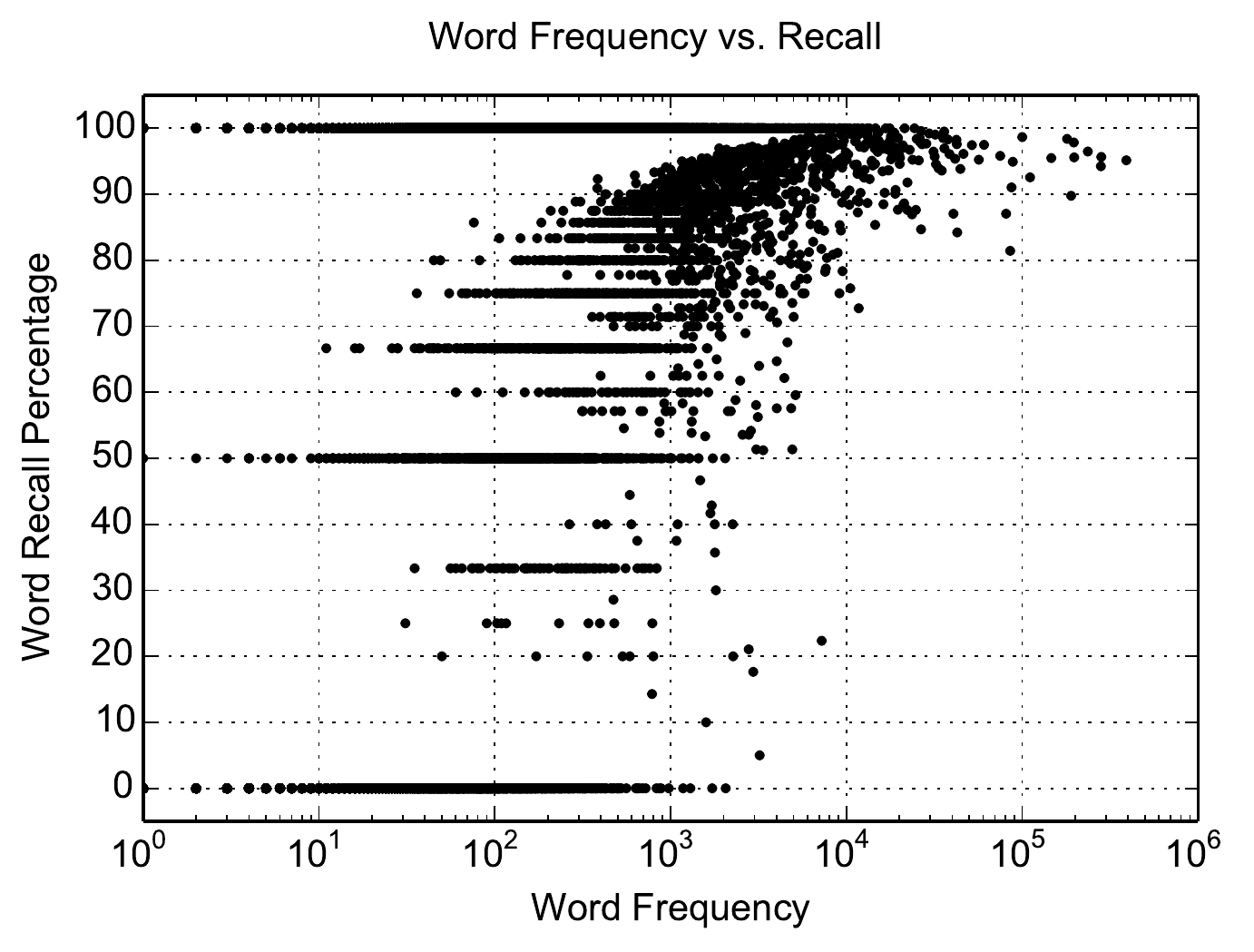} \caption{\small The correlation between word
frequency in the training distribution and recall in the test distribution. In general, rare words report worse recall compared to more frequent words.}
\label{fig:wordfreq} \end{figure}

We study the performance of our model on rare words. We use the recall metric
to indicate whether a word appears in the utterance regardless of position
(higher is better).  Figure \ref{fig:wordfreq} reports the recall of each word
in the test distribution as a function of the word frequency in the training
distribution.  Rare words have higher variance and lower recall while more
frequent words typically have higher recall. The word ``and'' occurs 85k times
in the training set, however it has a recall of only 80\% even after language
model rescoring.  The word ``and'' is frequently mis-transcribed as ``in''
(which has 95\% recall). This suggests improvements are needed in the language
model. By contrast, the word ``walkerville'' occurs just once in the training
set but it has a recall of 100\%. This suggests that the recall for a word
depends both on its frequency in the training set and its acoustic uniqueness.

\subsection{Interesting Decoding Examples}
\label{sec:int_examples}
In this section, we show the outputs of the model on several
utterances to demonstrate the capabilities of LAS.  All the
results in this section are decoded without a dictionary or a language
model.

During our experiments, we observed that LAS can learn multiple
spelling variants given the same acoustics. Table~\ref{tab:aaa} shows
top beams for the utterance that includes ``triple a''. As can be
seen, the model produces both ``triple a'' and ``aaa'' within the top
four beams. The decoder is able to generate such varied parses,
because the next step prediction model makes no assumptions on the
probability distribution by using the chain rule decomposition.  It
would be difficult to produce such differing transcripts using CTC due
to the conditional independence assumptions, where $p(y_i | x)$ is
conditionally independent of $p(y_{i + 1} | x)$. Conventional DNN-HMM
systems would require both spellings to be in the pronunciation
dictionary to generate both spelling permutations.



\begin{table*}[h!]
\centering
\caption{\small Example 1: ``triple a'' vs. ``aaa'' spelling variants.}
\label{tab:aaa}
\begin{tabularx}{\textwidth}{|l|X|r|r|}
\hline
\bfseries Beam & \bfseries Text & \bfseries Log Probability & \bfseries WER \\
\hline
\hline
Truth & call aaa roadside assistance & - & - \\
\hline
1 & call aaa roadside assistance & -0.5740 & 0.00 \\
2 & call triple a roadside assistance & -1.5399 & 50.00 \\
3 & call trip way roadside assistance & -3.5012 & 50.00 \\
4 & call xxx roadside assistance & -4.4375 & 25.00 \\
\hline
\end{tabularx}
\end{table*}

It can also be seen that the model produced ``xxx'' even though acoustically
``x'' is very different from ``a'' - this is presumably because the language
model overpowers the acoustic signal in this case. In the training corpus
``xxx'' is a very common phrase and we suspect the language model implicit
in the speller learns to associate ``triple'' with ``xxx''. We
note that ``triple a'' occurs 4 times in the training distribution and ``aaa''
(when pronounced ``triple a'' rather than ``a''-``a''-``a'') occurs only
once in the training distribution.

We are also surprised that the model is capable of handling utterances
with repeated words despite the fact that it uses content-based
attention. Table~\ref{tab:seven} shows an example of an utterance with
a repeated word.  Since LAS implements content-based attention,
it is expected it to ``lose its attention'' during the decoding steps
and produce a word more or less times than the number of times the
word was spoken. As can be seen from this example, even though
``seven'' is repeated three times, the model successfully outputs
``seven'' three times. This hints that location-based priors (e.g.,
location based attention or location based regularization) may not be
needed for repeated contents.


\begin{table*}[h!]
\centering
\caption{\small Example 2: Repeated ``seven''s.}
\label{tab:seven}
\begin{tabularx}{\textwidth}{|l|X|r|r|}
\hline
\bfseries Beam & \bfseries Text & \bfseries Log Probability & \bfseries WER \\
\hline
\hline
Truth & eight nine four minus seven seven seven & - & - \\
\hline
1 & eight nine four minus seven seven seven & -0.2145 & 0.00 \\
2 & eight nine four nine seven seven seven & -1.9071 & 14.29 \\
3 & eight nine four minus seven seventy seven & -4.7316 & 14.29 \\
4 & eight nine four nine s seven seven seven & -5.1252 & 28.57 \\
\hline
\end{tabularx}
\end{table*}

\section{Conclusions}
We have presented Listen, Attend and Spell (LAS), an attention-based neural
network that can directly transcribe acoustic signals to characters. LAS is
based on the sequence to sequence framework with a pyramid structure in the
encoder that reduces the number of timesteps that the decoder has to attend to.
LAS is trained end-to-end and has two main components.  The first component,
the listener, is a pyramidal acoustic RNN encoder that transforms the input
sequence into a high level feature representation. The second component, the
speller, is an RNN decoder that attends to the high level features and spells
out the transcript one character at a time. Our system does not use the
concepts of phonemes, nor does it rely on pronunciation dictionaries or HMMs.
We bypass the conditional independence assumptions of CTC, and show how we can
learn an implicit language model that can generate multiple spelling variants
given the same acoustics. To further improve the results, we used samples from
the softmax classifier in the decoder as inputs to the next step prediction
during training. Finally, we showed how a language model trained on additional
text can be used to rerank our top hypotheses.

\section*{Acknowledgements}
We thank Tara Sainath, Babak Damavandi for helping us with the data, language
models and for helpful comments. We also thank Andrew Dai, Ashish Agarwal, Samy
Bengio, Eugene Brevdo, Greg Corrado, Andrew Dai, Jeff Dean, Rajat Monga,
Christopher Olah, Mike Schuster, Noam Shazeer, Ilya Sutskever, Vincent
Vanhoucke and the Google Brain team for helpful comments, suggestions and
technical assistance.

\bibliographystyle{unsrt}
\bibliography{cites}

\appendix
\clearpage
\section{Alignment Examples}
In this section, we give additional visualization examples of our model and the attention distribution.

\begin{figure}[h!]
  \centering
  \includegraphics[width=4.5in]{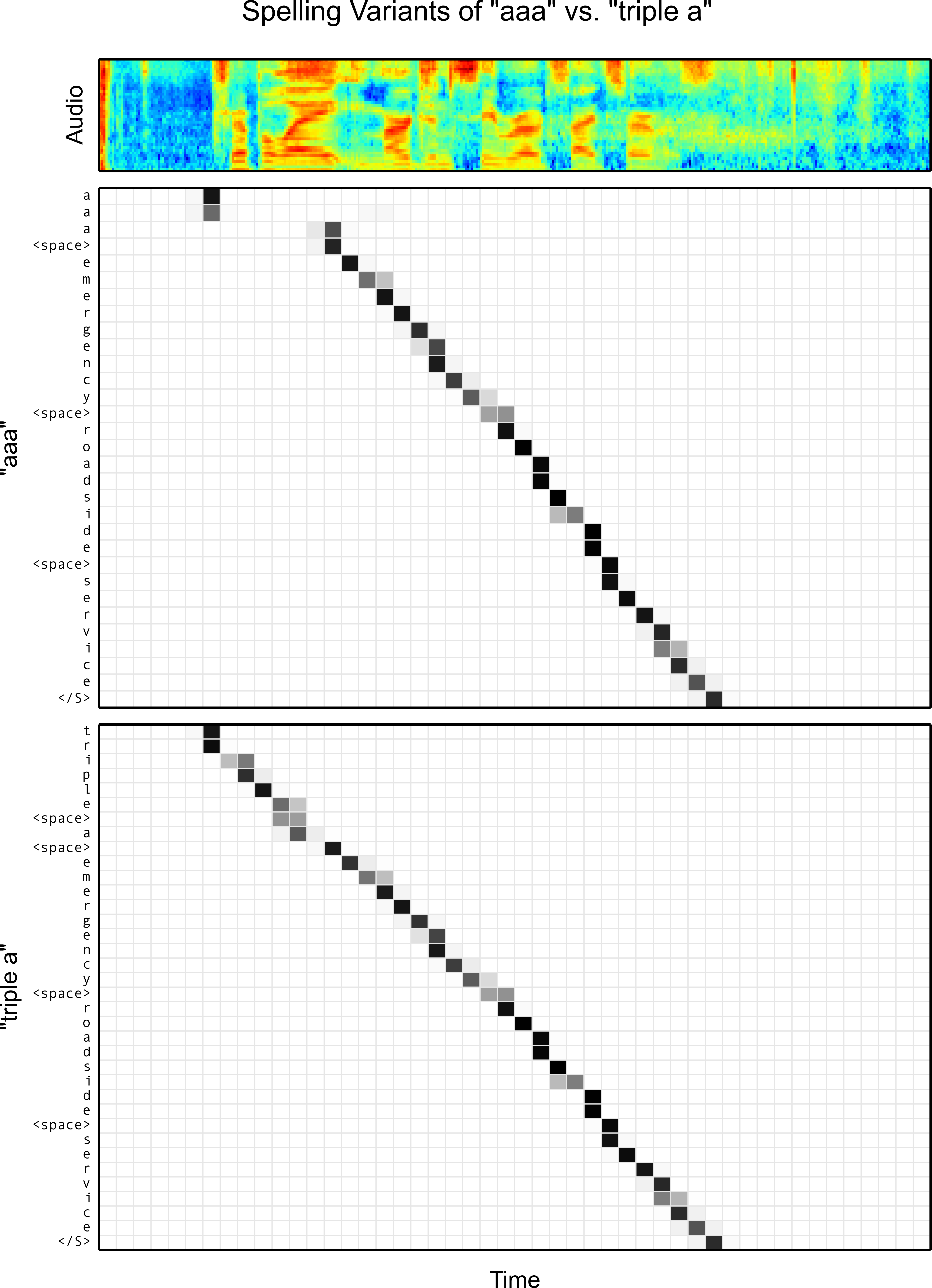}
  \caption{\small The spelling variants of ``aaa'' vs ``triple a'' produces different attention distributions, both spelling variants appear in our top beams. The ground truth is: ``aaa emergency roadside service''.}
  \label{fig:saint}
\end{figure}

\begin{figure}[h!]
  \centering
  \includegraphics[width=4.5in]{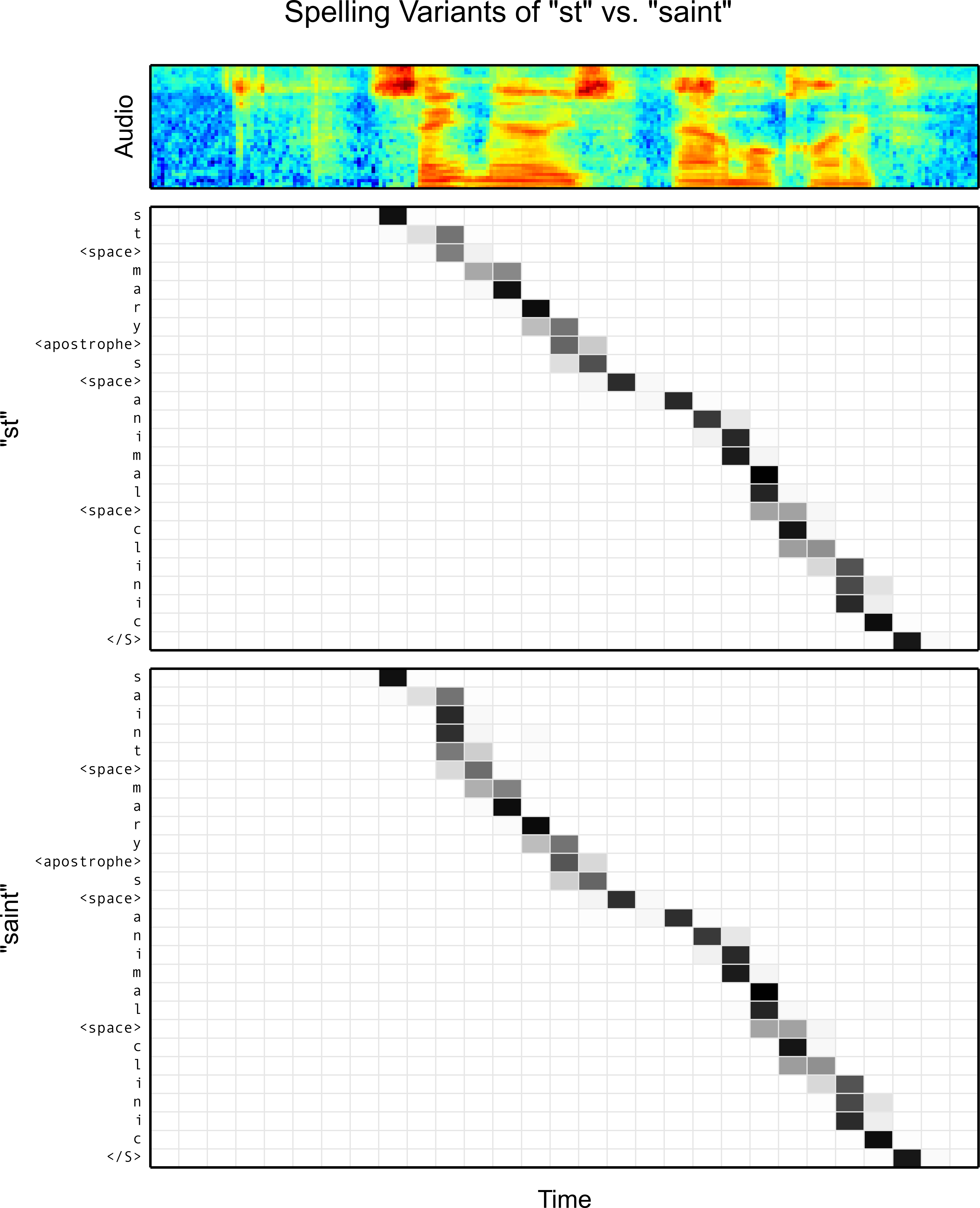}
  \caption{\small The spelling variants of ``st'' vs ``saint'' produces different attention distributions, both spelling variants appear in our top beams. The ground truth is: ``st mary's animal clinic''.}
  \label{fig:aaa.png}
\end{figure}

\begin{figure}[h!]
  \centering
  \includegraphics[width=4.5in]{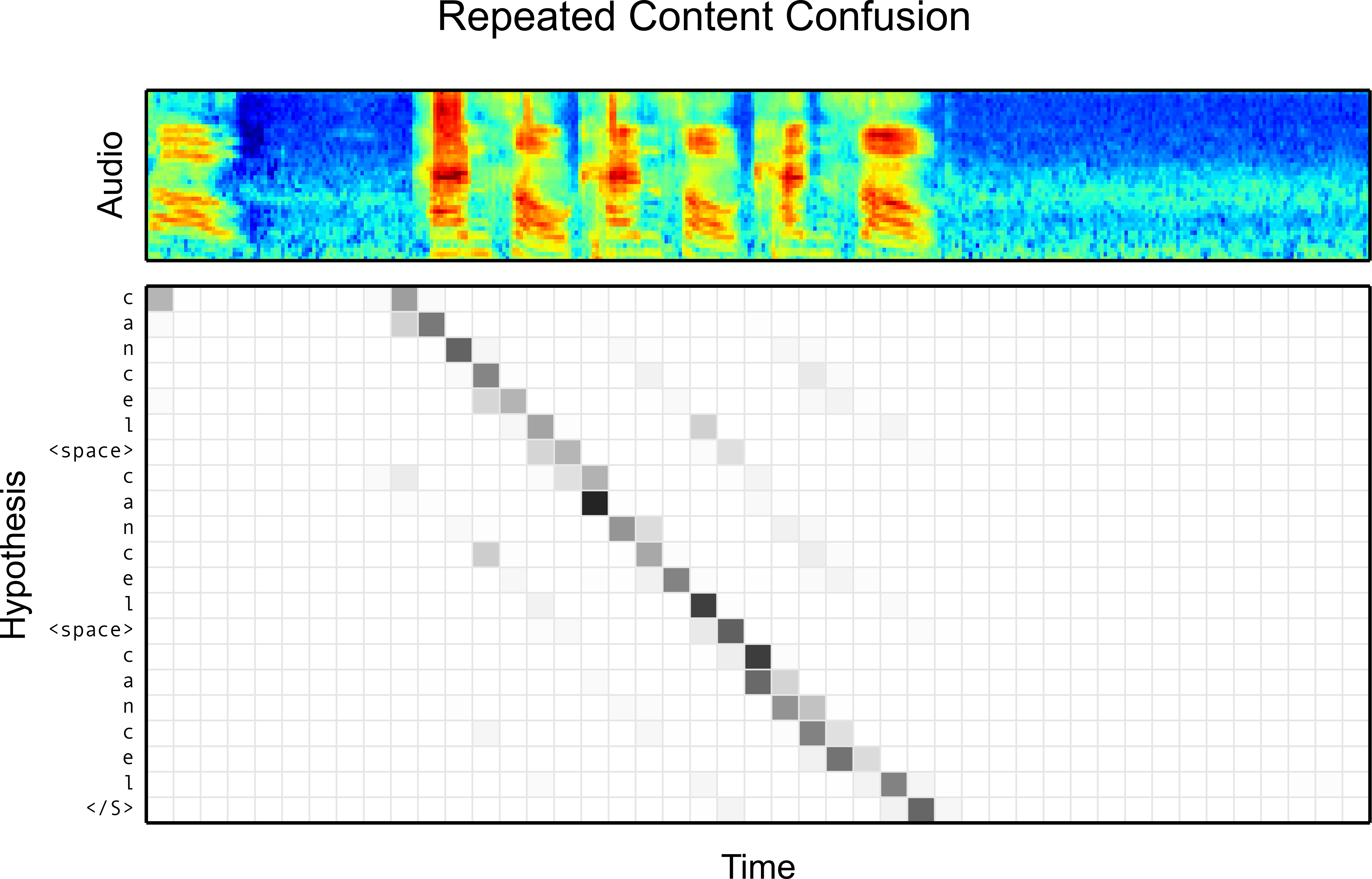}
  \caption{\small The phrase ``cancel'' is repeated three times. Note the parallel diagonals, the content attention mechanism gets slightly confused however the model still emits the correct hypothesis. The ground truth is: ``cancel cancel cancel''.}
  \label{fig:cancel.png}
\end{figure}



\end{document}